\documentclass{article} 
\usepackage[preprint]{colm2025_conference}
\usepackage{booktabs}
\usepackage{multirow} 

\usepackage{makecell}
\usepackage{soul}
\usepackage{stfloats}
\usepackage{color}
\usepackage{xcolor}
\definecolor{darkgreen}{RGB}{83,129,53}
\definecolor{darkred}{RGB}{163,21,21}

\usepackage{tikz}
\usepackage{subfig}
\usepackage{tcolorbox}
\usepackage{pgfplots}
\usepackage{wrapfig}
\usepackage{amssymb,mathrsfs}
\usepackage{amssymb}
\usepackage{array}
\usepackage{pgfplotstable}
\usepackage{pgf}
\usepackage[normalem]{ulem}
\usepackage{colortbl}
\usepackage{ulem}
\usepackage{xspace}

\usepackage{listings}

\usepackage{scalerel} 
\usepackage{times}
\usepackage{latexsym}
\usepackage{graphicx}
\usepackage{amsmath}
\usepackage{multirow}
\usepackage{amsthm}
\usepackage{amsfonts}
\usepackage{bm}
\usepackage{booktabs}
\usepackage{algorithm}
\usepackage{algpseudocode}
\usepackage{verbatim}
\usepackage{tabularx}

\usepackage{times}
\usepackage{latexsym}

\usepackage[T1]{fontenc}

\usepackage[utf8]{inputenc}

\usepackage{microtype}

\usepackage{inconsolata}

\usepackage{graphicx}

\usepackage{multirow}
\usepackage{booktabs}
\usepackage{adjustbox}

\usepackage{booktabs}   
\usepackage{threeparttable} 
\usepackage{graphicx}   
\usepackage{xcolor}     

\usepackage{amssymb}
\usepackage{microtype}
\usepackage{hyperref}
\usepackage{adjustbox}
\usepackage{color}
\usepackage{xcolor}
\usepackage{tcolorbox}
\usepackage{colortbl}
\usepackage{multicol}
\usepackage{url}
\usepackage{booktabs}
\usepackage{amsmath}
\usepackage{enumitem}
\usepackage{graphicx}
\usepackage{lineno}
\usepackage{xspace}
\usepackage{algorithm}
\usepackage{algpseudocode}
\usepackage{array}
\usepackage{booktabs}
\usepackage{longtable}
\usepackage{arydshln}
\usepackage{caption}

\definecolor{darkblue}{rgb}{0, 0, 0.5}
\hypersetup{colorlinks=true, citecolor=darkblue, linkcolor=darkblue, urlcolor=darkblue}

\title{\centering \raisebox{-.25\height}{\includegraphics[width=0.7cm]{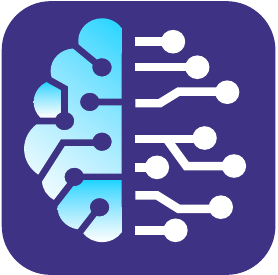}} KnowCoder-V2: Deep Knowledge Analysis}

\author{Zixuan Li$^{1,2}$\thanks{Co-first authors.}\ \ \thanks{Corresponding authors},  Wenxuan Liu$^{1,2,3}$\footnotemark[1], 
Long Bai$^{1,2}$\footnotemark[1], Chunmao Zhang$^{1,2}$\footnotemark[1], Wei Li, \\
\textbf{Fenghui Zhang}$^{1,2}$,  
\textbf{Quanxin Jin$^{1,2}$, }
\textbf{Ruoyun He$^{1,2}$, }
\textbf{Zhuo Chen$^{2,3}$, } 
\textbf{Zhilei Hu$^{1,2,3}$, } 
\textbf{Fei Wang$^{2}$, } \\
\textbf{Bingbing Xu$^{2}$, }   
\textbf{Xuhui Jiang$^{4,5}$, }   
\textbf{Xiaolong Jin$^{1,2,3}$\footnotemark[2], } 
\textbf{Jiafeng Guo$^{1,2,3}$\footnotemark[2], } 
\textbf{Xueqi Cheng$^{1,2,3}$} \\
$^1$Key Laboratory of Network Data Science and Technology, \\Institute of Computing Technology, Chinese Academy of Sciences \\
$^2$State Key Laboratory of AI Safety \\
$^3$School of Computer Science, University of Chinese Academy of Sciences \\
$^4$DataArc Tech Ltd, $^5$IDEA Research, International Digital Economy Academy \\
\texttt{\{lizixuan@ict.ac.cn, jinxiaolong@ict.ac.cn, guojiafeng@ict.ac.cn\}}
}

\newcommand{\KCI}{\textsc{KnowCoder-V1}\xspace}
\newcommand{\KCII}{\textsc{KnowCoder-V2}\xspace}
\definecolor{cvprblue}{rgb}{0.21,0.49,0.74}

\begin{document}



\maketitle



\begin{abstract}
Deep knowledge analysis tasks always involve the systematic extraction and
association of knowledge from large volumes of data, followed by logical
reasoning to discover insights. However, to solve such complex tasks, existing
deep research frameworks face three major challenges: 1) Coarse knowledge
management: They lack systematic organization and management of knowledge; 2)
Inefficient Operation Manner: They operate purely online, making it inefficient
for tasks that rely on shared and large-scale knowledge; 3) Shallow Knowledge
Computation: They cannot perform complex knowledge computation, limiting their
abilities to produce insightful analytical results. Motivated by these, in this
paper, we propose a \textbf{K}nowledgeable \textbf{D}eep \textbf{R}esearch
(\textbf{KDR}) framework that empowers deep research with deep knowledge
analysis capability. Specifically, it introduces an independent knowledge
organization phase to preprocess large-scale, domain-relevant data into
systematic knowledge offline. Based on this knowledge, it extends deep research
with an additional kind of reasoning steps that perform complex knowledge
computation in an online manner. To enhance the abilities of LLMs to solve
knowledge analysis tasks in the above framework, we further introduce
\textbf{\KCII}, an LLM that bridges knowledge organization and reasoning via
unified code generation. For knowledge organization, it generates instantiation
code for predefined classes, transforming data into knowledge objects. For
knowledge computation, it generates analysis code and executes on the above
knowledge objects to obtain deep analysis results. Experimental results on more than thirty datasets across six knowledge analysis tasks demonstrate the effectiveness of \KCII. Moreover, when integrated into the KDR framework, \KCII can generate high-quality reports with insightful analytical results compared to the mainstream deep research framework.
\renewcommand{\thefootnote}{}%
\footnotetext{Preprint. Work in progress.}%
\renewcommand{\thefootnote}{\arabic{footnote}}%

\end{abstract}

\section{Introduction}

Recently, Large Language Models (LLMs) have demonstrated remarkable capabilities
in natural language understanding and generation~\citep{zhaosurvey}. Through
inference-time scaling, some LLMs, such as
OpenAI-o1~\footnote{https://openai.com/index/openai-o1-system-card/} and
DeepSeek-R1~\citep{guo2025deepseek}, allocate increased computational resources
to generate longer chains of thought before producing responses, resulting in
improved reasoning performance~\citep{snell2024scaling}. These advancements have
given rise to deep research, a new category of abilities to autonomously tackle
complex tasks by searching and synthesizing information from diverse online
sources~\citep{Li2025webthinker}. 

\begin{figure*}
   \centering
   \includegraphics[width=\textwidth]{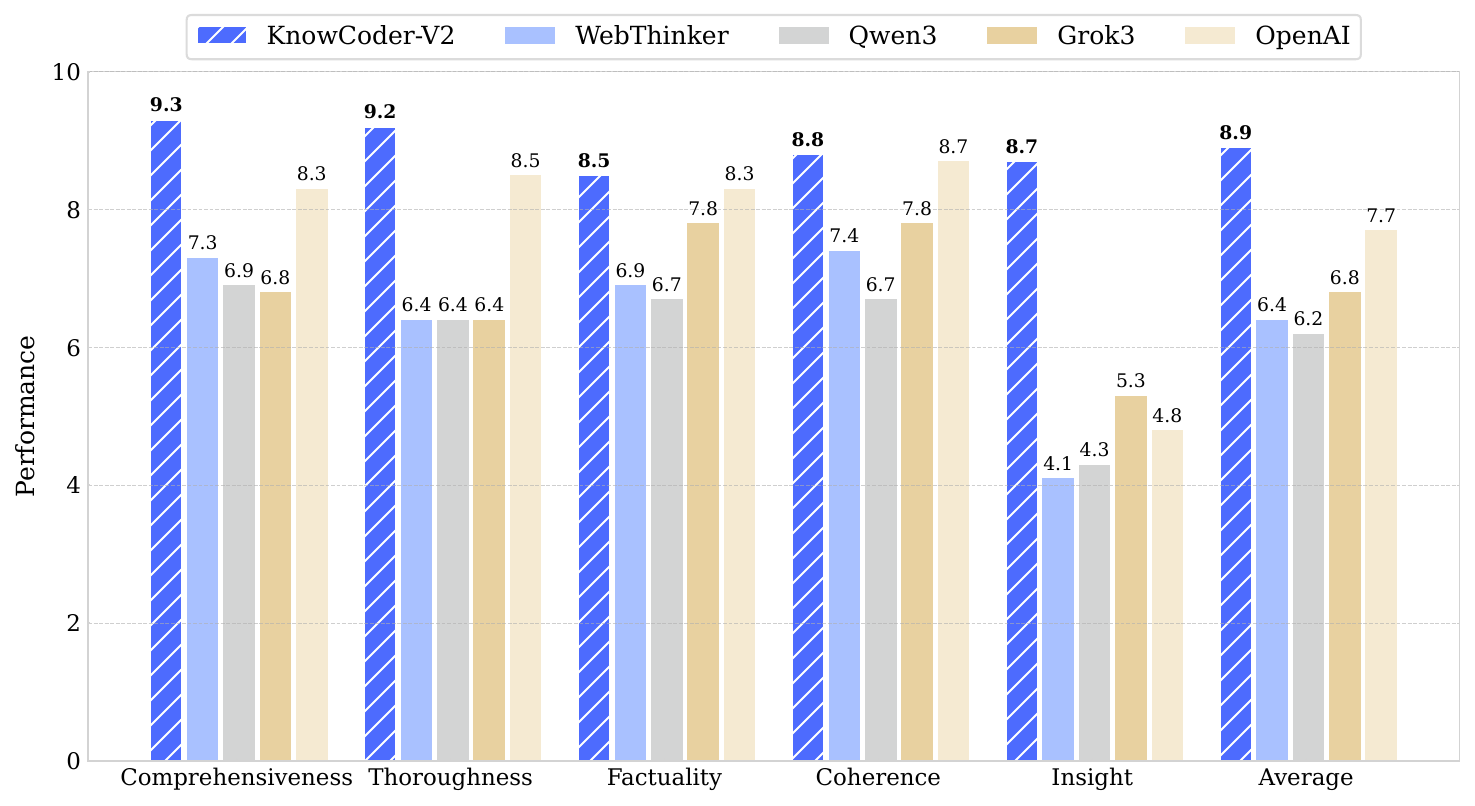}
   \caption{Report generation performance of the Knowledgeable Deep Research framework empowered by KnowCoder-V2 and its counterparts. Scores are averaged from DeepSeek-R1 and DeepSeek-V3 evaluations.}
   \label{fig:report_gen}
\end{figure*}

Although deep research framework shows considerable capability to handle several
kinds of complex tasks, it falls short when dealing with one particular kind of
tasks, which we refer to as deep knowledge analysis task. These tasks involve
the systematic extraction, organization, and association of knowledge from
large-scale and heterogeneous data sources, followed by multi-step logical
reasoning over the such structured knowledge. For instance, the problem ``What
were the most influential research contributions to quantum computing between
2020 and 2024?'' cannot be addressed through simple web search and text
generation, especially when the analyses about the problem are not publicly
available. Specifically, the limitations of the exising deep research framework
can be categorized into three key dimensions:

1) \textbf{Coarse Knowledge Management}: The existing deep research framework
search limited web pages to look up for information related to the task. In such
frameworks, the basic unit of external knowledge is the web page. This results
in coarse knowledge management, which means the models are hard to exploit
fine-grained and organized knowledge distributed in thousands of documents.

2) \textbf{Inefficient Operation Manner}: The existing deep research framework
operate entirely online, which may repeatedly searches and processes similar web
pages across different tasks. This operation manner involves unnecessary
computational overhead and hinders the efficiency of the framework.

3) \textbf{Shallow Knowledge Computation}: The existing deep research framework
process text-form knowledge via text generation, which is only able to perform
shallow knowledge computation. In contrast, deep knowledge analysis requires
more sophisticated computational capability of knowledge, including aggregating
information across entities and timestamps, performing logical deduction and
statistical inference, and dynamically querying and transforming structured
data. Without these capabilities, the deep research framework falls short of the
depth and precision needed to support insightful analysis.

Motivated by these, in this paper, we proposes a Knowledgeable Deep Research
(KDR) framework that augments existing deep research systems with deep knowledge
analysis capabilities. KDR contains two phases, i.e., knowledge organization and
knowledge reasoning phase. In the knowledge organization phase, large-scale,
domain-specific, and multi-source data are systematically preprocessed into
structured knowledge based on a task-specific ontology. After this phase, we can
obtain a structured knowledge base. In the knowledge reasoning phase, based on
this organized knowledge base, KDR conducts complex knowledge computations in
real time. Specifically, KDR further calls two kinds of reasoning steps to solve
each subtask autonomously, i.e., deep computing step and deep search step. The
former aims to search for the structured knowledge and conduct complex
computation to obtain the analysis results. The latter aims to search for the
online website to get the related documents. With the two kinds of materials as
input, KDR writes the final report. 

To address the limitations of current LLMs in deep knowledge analysis tasks, we
propose \KCII, an LLM that unifies knowledge organization and reasoning through
code generation. In the knowledge organization phase, \KCII generates
instantiation code for predefined classes, systematically converting raw data
into structured knowledge objects. In the reasoning phase, it produces
analytical code that operates on these objects to perform complex computations
and multi-step analysis. Experimental results on thirty datasets across six types of knowledge analysis tasks, including ontology expansion, knowledge extraction, and knowledge graph question answering, demonstrate the effectiveness of the proposed \KCII. Furthermore, by integrating \KCII, the proposed KDR framework can generate high-quality reports with comprehensive experimental analyses and deep insights.

In summary, our key contributions are from four aspects:
\begin{itemize}[leftmargin=*]
    \item We propose a knowledgeable deep research framework that empowers the
 existing deep research framework with deep knowledge analysis.
    \item To enhance the abilities of LLMs to solve the knowledge analysis tasks
 in the above framework, we introduce \KCII, an LLM that bridges knowledge
 organization and reasoning via unified code generation. 
    \item Experimental results on more than thirty datasets across six knowledge
 analysis tasks demonstrate the effectiveness of the proposed \KCII.
    \item With \KCII, the proposed knowledgeable deep research framework can
 produce high-quality reports with deep analysis results.
\end{itemize}

In the following sections, we will first introduce the knowledgable deep
research framework in Section~\ref{sec:framework}. Then, we will present the
proposed \KCII in Section~\ref{sec:kc2}. After that, we will evaluate the
performance of \KCII and the proposed framework in Section~\ref{sec:experiment}.
Finally, we will conclude the paper in Section~\ref{sec:conclusion}.

\section{Related Work}

\subsection{LLM-based Knowledge Organization}

For the knowledge organization phase, the related work primarily consists of
three key tasks, i.e., ontology expansion, knowledge extraction, and knowledge
update. Ontology expansion aims to involve integrating emerging new concepts
into existing ontology structures by identifying their appropriate parents.
Early works~\citep{takeoka2021low, shen2024unified} fine-tuned BERT-based models
to leverage textual descriptions of concepts and transfer this task into a
multi-class classification task. Recently, some
works~\citep{moskvoretskii2024taxollama,zeng2024codetaxo} use LLMs to conduct
this task and get significant improvement. Knowledge Extraction refers to
leveraging LLMs to extract structured knowledge following the given concepts
from the ontology, which can be divided into in-context learning (ICL)-based and
supervised-based extraction. The ICL-based extraction
\cite{wang2023code4structcodegenerationfewshot,li2023codeielargecodegeneration,guo2023retrievalaugmentedcodegenerationuniversal}
aims to leverage the universal capabilities of LLMs to extract the knowledge
directly without fine-tuning the specific datasets. And Supervised-based
Extraction~\cite{sainz2024gollieannotationguidelinesimprove,li2024knowcoder,zuo2024alignxie}
aims to construct a large IE corpus to fine-tune the LLM to better adapt to more
specific domains or large-scale schema scenarios. Knowledge update refers to
updating the out-of-date knowledge according to the given new extracted
knowledge. The related techniques include entity linking, entity alignment, and
so on.

\subsection{LLM-based Knowledge Reasoning}
The techniques related to the knowledge reasoning phase include
Retrieval-Augmented Generation (RAG), Graph Retrieval-Augmented Generation
(GraphRAG), Deep Research, and so on. RAG mitigates hallucinations by retrieving
relevant documents as context for the LLM~\citep{xu2024activerag, asai2023self,
shao2023enhancing}. GraphRAG extends this by extracting knowledge graphs from
text and constructing community hierarchies, thereby improving the LLM's
comprehension of complex datasets~\citep{edge2024local, ma2024think}.

Building on RAG, Deep Search or Deep Research further focuses on in-depth,
multi-step investigation and analysis of information from diverse sources for
deeper insights. This capability is commonly integrated as an advanced search
feature in proprietary LLMs such as Sonar Reasoning Pro, and GPT-4o Search
Preview. Works on open-source LLMs focus on augmenting reasoning for deep
search. For instance, Search-o1, ODS~\citep{li2025search, alzubi2025open}
augment the latest open-source reasoning LLMs by injecting the capabilities of
web search tools to answer queries; others~\citep{jin2025search, song2025r1,
chen2025learning} train LLMs with reinforcement learning to autonomously
generate search queries during step-by-step reasoning. Besides,
WebThinker~\citep{Li2025webthinker} employs LLMs to autonomously search, explore
web pages deeply, and draft research reports within its reasoning process.
Agentic Reasoning~\cite {wu2025agentic} enhances LLMs by integrating agents that
utilize external tools and construct knowledge graphs for deductive reasoning.
DeepResearcher~\citep{zheng2025deepresearcher} focuses on training LLM agents in
realistic web environments to perform iterative reasoning and search.
\section{Knowledgeable Deep Research}\label{sec:framework}

\begin{figure*}
    \centering
    \includegraphics[width=\textwidth]{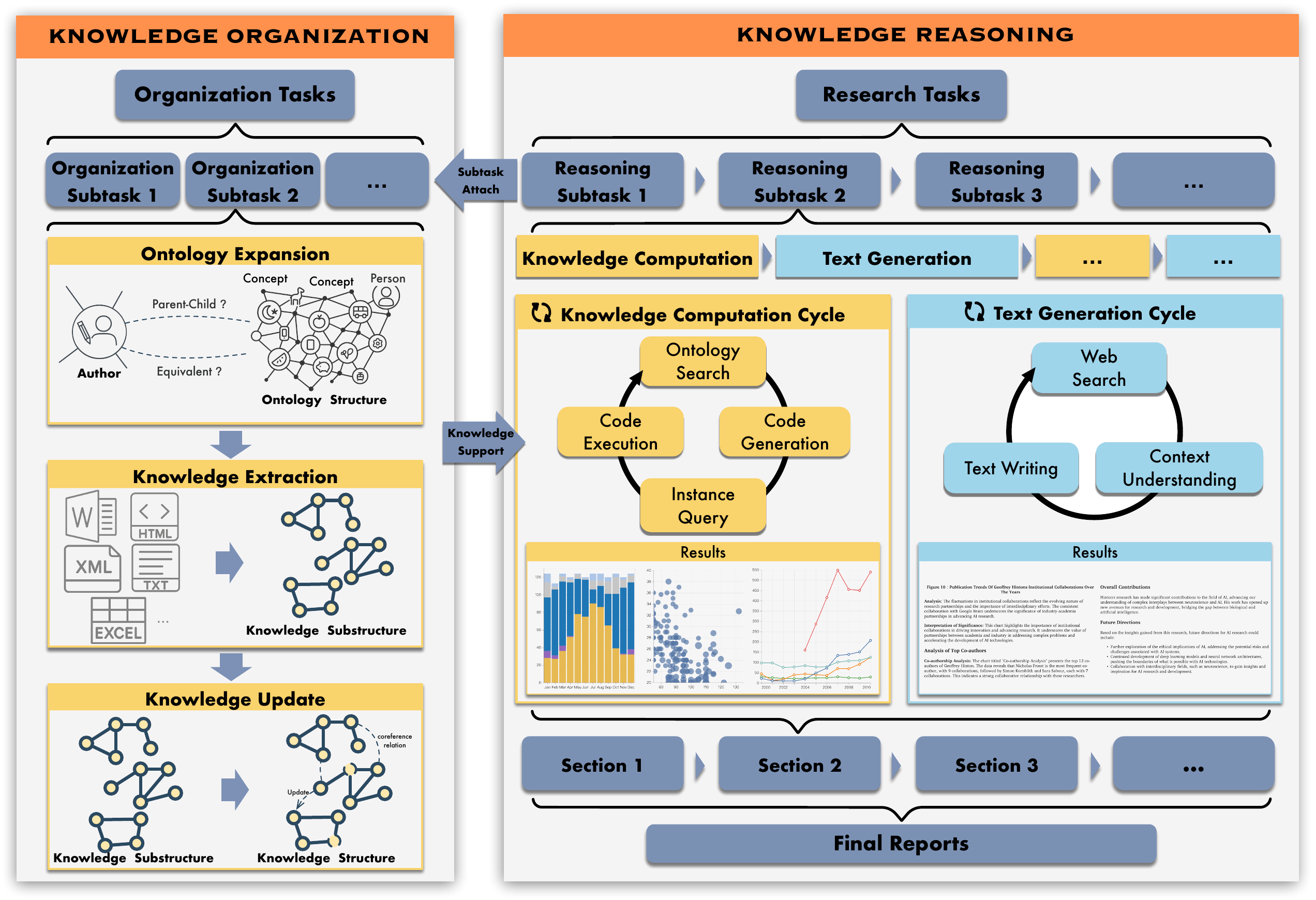}
    \caption{Demonstration of the proposed knowledgeable deep research
    framework. The mustard yellow blocks indicate tasks related to deep
    knowledge analysis, which includes knowledge organization and computation
    tasks. }
    \label{fig:main}
\end{figure*}

In the following sections, we introduce the proposed Knowledgeable Deep Research
(KDR) framework. As demonstrated in Figure~\ref{fig:main}, KDR distinguishes
itself from traditional deep research framework by clearly separating knowledge
organization into a dedicated offline phase. This decoupling allows the
structured organization of vast amounts of research objects into accessible
structured knowledge. Leveraging these structured knowledge, the KDR framework
then enables rapid and in-depth analysis of research objects during the
subsequent knowledge reasoning phase.



\subsection{Knowledge Organization Phase}

As shown in the left part of Figure~\ref{fig:main}, given a research topic as
input, the offline knowledge organization phase first generates several
organization subtasks, where each subtask corresponds to a concept that needs to
be extracted. Then, for each subtask, this phase consists of three steps, i.e.,
ontology expansion, knowledge extraction, and knowledge update. First, in the
ontology expansion step, KDR aligns the focused concepts to the existing
ontology structure. Next, the knowledge extraction step transfers a large
collection of related information into the corresponding structured knowledge.
Finally, the knowledge update step updates the newly extracted knowledge into
existing knowledge bases.

\paragraph{Task Decomposition.} Given a research topic as input, the offline
knowledge organization process first decomposes the overall task into several
subtasks, each corresponding to specific research concepts such as entities and
events. Entity concepts are associated with particular properties and relation
types, while event concepts have defined argument types. For instance,
considering the research question, "Please help me research the papers published
by Geoffrey Hinton in the past decade and analyze the trends in his research
interests over time", the related research entities would include academic
papers along with properties such as publication date, authorship, citation and
so on. Due to the critical role of this step, the process always employs a
"human-in-the-loop" strategy: initially, LLMs propose the structures of
concepts, which are subsequently reviewed and validated by human experts.


\paragraph{Ontology Expansion.} As new organizational tasks emerge, concepts
defined in these subtasks often have different labels but represent the same
underlying meanings, or they may have associations with the existing concepts in
the ontology. For instance, "author" is a subclass of "person" and is equivalent
to the concept of "authors." To facilitate subsequent fusion of structured
knowledge, the ontology expansion step systematically associates the new
concepts with the existing ontology. Starting from the root concepts "Entity"
and "Event," this step involves identifying both equivalence and hierarchical
(hyponym) relationships among the concepts, ensuring consistency and semantic
coherence within the ontology.

\paragraph{Knowledge Extraction.}\label{sec:framework-extraction} With the ontology as
input, the extraction step processes various types of documents, including PDF,
TXT, WORD, and HTML formats, to extract corresponding entities and events. This
step can be formulated as the Information Extraction (IE)
tasks~\cite{cowie1996information}, specifically designed to derive structured
knowledge from unstructured data. Typical IE subtasks performed here include
Named Entity Recognition (NER), Relation Extraction (RE), and Event Extraction
(EE).

\paragraph{Knowledge Update.}\label{sec:rollout} Since knowledge is often
dispersed across multiple documents, extracting information from a single
document or text snippet usually provides only partial details about an entity
or event. For example, assembling comprehensive data on a specific author's
publications requires extracting relevant information from multiple sources,
such as all their published papers. Additionally, knowledge can evolve over
time. For instance, a person's residence might change. This step continuously
updates the new extracted knowledge to previously extracted ones to ensure
ongoing accuracy and consistency.

\subsection{Knowledge Reasoning Phase}

Given a specific research task, the knowledge reasoning phase generates a
comprehensive report based on previously organized knowledge. As illustrated on
the right side of Figure~\ref{fig:main}, this phase begins by decomposing the
overall research task into multiple subtasks, each corresponding to a specific
part of the final report. Each subtask is addressed through two types of
reasoning steps, i.e., knowledge computation and text generation. Using the
outputs from these reasoning steps, the complete and coherent research report is
ultimately obtained.

\paragraph{Task Decomposition.} This step decomposes the overall research task
into several smaller reasoning subtasks. For each subtask, LLMs further
adaptively carry out two distinct reasoning steps, i.e., knowledge computation
and text generation, by generating the corresponding queries. Specifically, the
knowledge computation step retrieves structured knowledge from the knowledge
organization phase and generate codes to conduct necessary computational
analyses or experiments. The text generation step, on the other hand, involves
retrieving and synthesizing information from online sources to support content
generation.

\paragraph{Knowledge Computation Cycle.} 

The knowledge computation cycle is automatically triggered when the
corresponding queries are generated during the task decomposition step. Since
knowledge computation involves large-scale data and thus cannot be included
directly in the prompt, this reasoning step aims to generate the computation
code based on the data ontology, rather than directly accessing the data.
Specifically, it follows a cyclic process. In the ontology search step, it
searches for the structures of relevant concepts and represents them in formats
such as JSON, text, or code. Then, in the code generation step, code is
generated based on the identified ontology structure. Following that, the
instance query step retrieves instance data based on the ontology structure
defined earlier. Finally, in the code execution step, the analysis code is
executed on the retrieved instance data to generate the analysis results. If the
code execution fails or the generated analysis results do not meet the required
criteria, this cycle will be repeated until the desired results are achieved or
the maximum number of iterations is reached.

\paragraph{Text Generation Cycle.} 
The text generation cycle is automatically triggered when the corresponding
queries are generated during the task decomposition step. This cycle searches
for relevant information on the internet and uses it for subsequent text
generation. Specifically, in the web search step, the system searches online
websites using the decomposed query as input. Then, the LLM processes the
retrieved websites and generates corresponding text. If the generated text does
not sufficiently address the query, this cycle will be repeated until the
desired results are achieved.

\paragraph{Merge and Revise.} 
After the above two kinds of reasoning steps, the materials for a subtask are
ready. Then, the merge and revise step will merge the materials to obtain the
section content. Then, it will finally merge each section to form the complete
report. Besides, an extra revision step is conducted to polish the report from a
global perspective.

\section{KnowCoder-V2}\label{sec:kc2}

Although LLMs have demonstrated strong capabilities in many natural language
processing tasks, they still face limitations when it comes to deep knowledge
analysis. To address these challenges, we introduce \KCII, an LLM designed to
bridge knowledge organization and reasoning through a unified code generation
approach. In the following sections, we present the details of how KDR framework
is implemented with \KCII.

\begin{figure*}[t]
    \centering
    \includegraphics[width=1.0\textwidth]{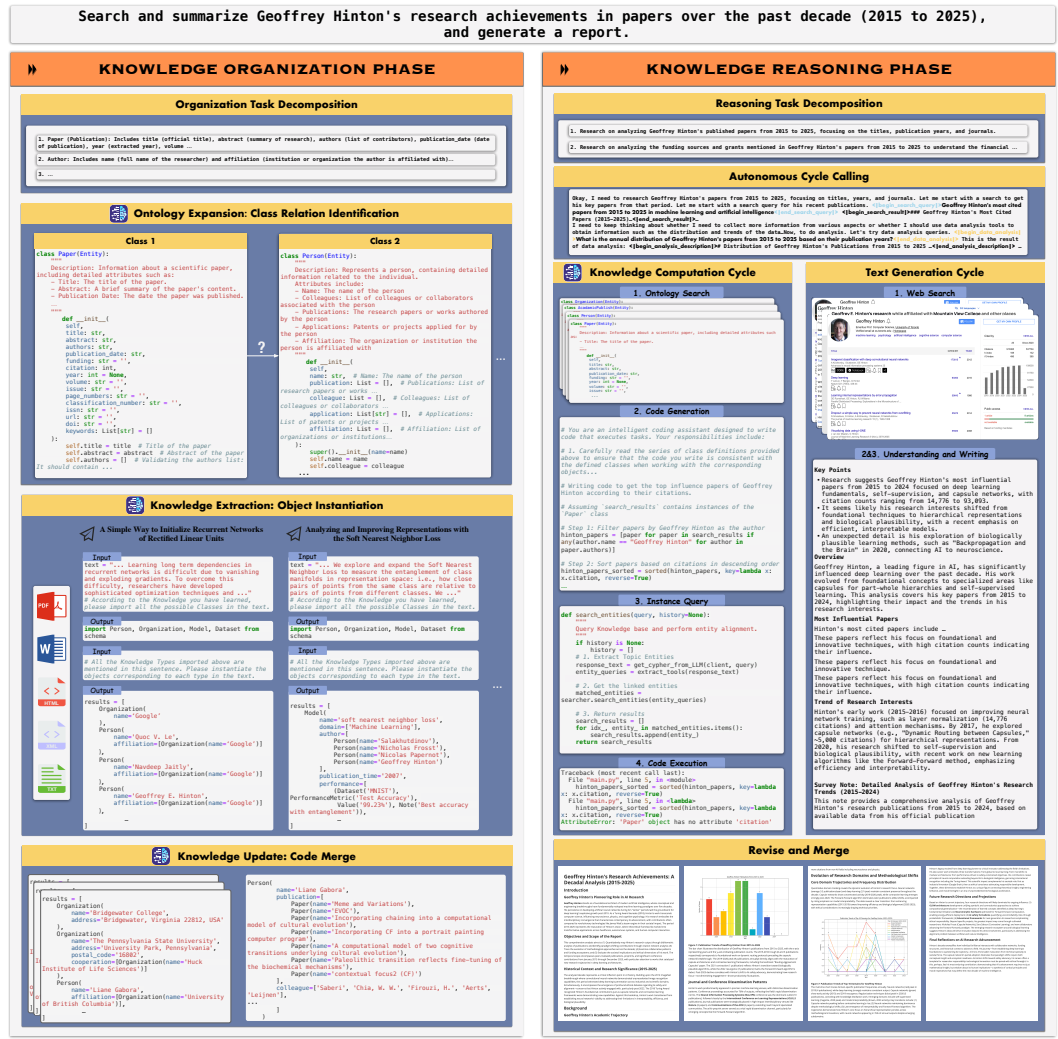}
    \caption{Illustration of the Knowledgeable Deep Research Framework implemented by \KCII. We use DeepSeek-R1 to handle the parts of the KDR framework that are not related to knowledge analysis tasks.}
    \label{fig:methodology}
\end{figure*}

\subsection{Knowledge Organization Phase}
The knowledge organization phase aims to systematically induce, extract, and
update the structure knowledge related to the given research task. Following
\KCI~\citep{li2024knowcoder}, we use Python classes to represent concepts in the
ontology and Python objects to represent knowledge instances. As illustrated in
the left side of Figure~\ref{fig:methodology}, two base classes, i.e., Entity
and Event, are predefined. All other domain-specific concepts are implemented as
subclasses of these base classes. For example,
\textcolor{darkgreen}{\texttt{class Person(Entity)}}. The attributes of each
concept (such as a person's affiliation) are defined as member variables within
the class. Additionally, examples and descriptions for each concept are provided
in the class docstrings, consistent with \KCI.

\subsubsection{Task Decomposition}
Given a research task, we begin by decomposing it into a set of subtasks using a
human-in-the-loop approach. First, the research task is provided as the input to
an LLM, which infers the relevant concepts and their associated attributes.
Next, these inferred concepts are manually reviewed to ensure accuracy and
completeness. Finally, the validated concepts are automatically structured into
Python classes.

\subsubsection{Ontology Alignment: Class Relation Identification}


The same concept may have different definitions across various knowledge
organization tasks, and hierarchical relationships may exist between distinct
concepts. This misalignment poses significant challenges for subsequent
knowledge organization and reasoning. To address this, the ontology expansion
step aims to identify semantic relations, including parent-child and equivalent
relations, by analyzing the definitions of the corresponding concepts. To
achieve this, we design a two-stage alignment process that combines pre-trianed
and large language models. In the first stage,
following~\cite{zeng2024codetaxo}, SimCSE~\cite{gao2021simcse} is adopted to
encode the concept names and descriptions into $m$-dimensional vectors. Based on
the cosine similarity among these vectors, we retrieve the top-$k$ most similar
candidate concepts. In the second stage, we refine the alignment results by
prompting LLM with the code definitions of the candidate classes to determine
whether they exhibit parent-child or equivalent relations.

\subsubsection{Knowledge Extraction: Object Instantiation}\label{sec:extraction}
In this step, \KCII extracts the corresponding structured knowledge by
instantiating the classes based on the given text. Previous
work~\citep{li2024knowcoder,wang2023instructuiemultitaskinstructiontuning}
requires including all class definition code within the prompt. This presents a
significant challenge: when a large number of concepts need to be extracted, the
total class definition code can become excessively long and may even exceed the
prompt length limit. As a result, both the effectiveness and efficiency of the
extraction process can degrade due to the excessive length of prompts.

To address this issue, existing methods adopt a partitioned extraction
strategy~\citep{gui2024iepile, liu2025towards}. They divide the concepts into
several groups and construct a separate prompt for each group, which are then
processed individually by the LLMs to extract the corresponding knowledge. While
this strategy can improve performance, it significantly reduces efficiency. In
contrast, elegant code design does not require developers to repeatedly provide
the full definition code for the same classes; instead, they can simply import
the predefined classes. Motivated by this, we internalize class definitions into
the parameters of \KCII, allowing it to use a simple import clause during the
extraction process. To support these new capabilities, we introduce two types of
training tasks, i.e., ontology understanding and following tasks to implicitly
enhance \KCII's ability to comprehend and operate on dynamic, large-scale
knowledge organization tasks. 

\paragraph{Ontology Understanding Tasks.}
To enable LLMs to understand the detailed definition of these classes, we define
a ontology understanding task. After importing a class, we instruct LLMs to
generate the detailed definitions with the prompt of ``\textit{Please generate
the detailed schema of the class \{type\} from \{task\} based on your
memory.}''. 

\paragraph{Ontology Following Tasks.}
Ontology following task aims to enhance the ability to follow the imported
classes and generate instantiating code accordingly. It consists of two
subtasks, i.e., class importing and object instantiate tasks. Given the text,
class importing task aims to recall the corresponding types via completing the
import clause. Given the text and import clause, object instantiate task aims to
extract corresponding knowledge via generating the instantiating code. In some
situations, users may predefine types that they want to extract but in other
situations, the user may not give out the types explicitly and want the model to
recall all the possible types in its memory. Thus, we conduct training under two
settings, i.e., close domain and open domain settings. In the following, we will
introduce details of the training data.

\begin{itemize}
    \item \textbf{Class Importing Data.} In the closed-domain setting, users
    specify a predefined set of classes to be extracted. In this scenario, we
    begin by explicitly listing the relevant types using the import clause:
    \textit{From \{task\} Import \{type1\}, \{type2\}}. Besides, we also give
    out the task description in the prompt: \textit{Some \{task\} Types are
    imported above. Please instantiate all the possible \{task\} Objects in the
    following sentence. }. Then, \KCII is instructed to recall the most
    potential types in the text. The output is the potential knowledge types in
    the form of \textit{Import \{type1\}}.

    In the open domain setting, we remove the given schema and only instruct LLM
    to recall possible types from knowledge types the model have learned by the
    task description: \textit{According to the \{task\} Types you have learned,
    please import all the possible \{task\} Types in the sentence}. The output
    is potential knowledge types in the form of \textit{Import \{type1\}}.
    \item \textbf{Object Instantiation Data.} After obtaining the potential
    knowledge types in the text, we aim to instantiate a concrete knowledge
    object by task description: \textit{Some \{task\} Types are imported above.
    Please instantiate all the possible \{task\} Objects in the following
    sentence. }. The output is a list of instantiated objects same with direct
    extraction data.
\end{itemize}

\subsubsection{Knowledge Update: Code Merge}  In this step, \KCII merges
entities and events that share the same name. For these entities or events,
their relations, properties, or arguments are combined into a single unified
entity or event. Additionally, if previously stored knowledge conflicts with
newly extracted knowledge, the system updates it with the latest knowledge. This
heuristic strategy proves effective in most cases. Developing more precise and
robust knowledge update mechanisms is left as future work.

\subsection{Knowledge Reasoning Phase}

In this section, we introduce the implementation details of the knowledge
reasoning phase with \KCII.

\subsubsection{Task Decomposition}

As illustrated on the right side of Figure~\ref{fig:methodology}, the knowledge
reasoning phase begins by decomposing the given research task into several
subtasks. Each subtask addresses a research topic closely related to the
research task. For each subtask, an LLM further decomposes it into two distinct
reasoning steps, i.e.,  the knowledge computation cycle and the text generation
cycle. Specifically, we use the tags
\textcolor{darkgreen}{\texttt{<begin\_data\_analysis>}} and
\textcolor{darkgreen}{\texttt{<end\_data\_analysis>}} to indicate the queries of
the knowledge computation cycle and the tags
\textcolor{darkgreen}{\texttt{<begin\_web\_search>}} and
\textcolor{darkgreen}{\texttt{<end\_web\_search>}} to indicate the queries of
the text generation cycle.

\subsubsection{Knowledge Computation Cycle}
Since various types of knowledge have already been encapsulated as distinct
Python objects during the knowledge organization phase, \KCII implements the
knowledge computation cycle to generate executable code based on the queries
derived from task decomposition. Specifically, this cycle produces three types
of code: class definition code, knowledge declaration code, and knowledge
computation code. By integrating and executing these components, \KCII is able
to perform deep and precise computations over the structured knowledge.

\paragraph{Ontology Search.}
Given a research subtask, the ontology search step aims to identify relevant
concepts from the ontology. With the queries as input, we leverage Elasticsearch
to retrieve code-based concepts that are most pertinent. Once identified, these
concepts are then transformed into their corresponding class definition code,
enabling seamless integration into subsequent knowledge computation processes.

\paragraph{Code Generation.}
Analysis tasks typically require large-scale data, making it impractical to
include all the necessary data directly in the prompt. More importantly, when
humans write code, they usually complete the entire script first and then input
the data. Inspired by this manner, we assume the relevant objects has already
been retrieved, instantiated, and stored in a variable named
\textcolor{darkgreen}{\texttt{search\_results}}. Consequently, \KCII directly
generates the code required to complete the corresponding queries based solely
on the class definition obtained from the ontology search step.

\paragraph{Instance Query.}
In this step, we instantiate the corresponding entity objects and store them in
the \textcolor{darkgreen}{\texttt{search\_results}} list. First, we perform
named entity recognition using an LLM to extract the topic entities of the
queries. Next, using these extracted entity names, we use elasticsearch and
entity names to query the knowledge base to find related entities. Finally,
based on the subgraphs related to each entity, we instantiate the corresponding
objects. Note that, if the   we query the knowledge base to find related entries. Finally, we
instantiate the corresponding entity objects and store them in the list named
\textcolor{darkgreen}{\texttt{search\_results}}.

\paragraph{Code Execution.}
In this step, we integrate the class definition code, the filed
\textcolor{darkgreen}{\texttt{search\_results}} variables, and the analysis code
to create the final executable script. By running this combined code in the
Python execution environment, we obtain the corresponding analysis results.
However, there could be situations where the code execution fails or produces
incorrect results (e.g., generating empty charts). To address this, we
incorporate an analysis evaluation agent that reviews the obtained analysis
results and determines whether they satisfy the task requirements. If the
results do not meet the specified criteria, the process loops back to the first
step for corrections.

\subsubsection{Text Generation Cycle}
With the search query input, the model iteratively conduct web search until the
query can be answered. Specifically, it contains three steps, i.e., web search,
context understanding, and text writing. As these steps are broadly implemented
in the existing deep research frameworks, we will not detail them in this paper.

\subsubsection{Merge and Revise}
A significant challenge is ensuring consistency between the results generated by
the knowledge computation and text generation cycles. Inconsistencies often
arise due to incomplete search results, quality issues in web content, or
inadequacies in the structured knowledge. To address this, we designed an
additional agent to merge results from both cycles and revise them. This agent
autonomously evaluates the analysis results from the knowledge computation cycle
alongside the generated results from the text generation cycle, identifies
conflicting content, and resolves these discrepancies to achieve consistency.

\subsection{Training of KnowCoder-V2}
To enhance the extraction capabilities of \KCII, we utilized 14 Named Entity
Recognition (NER) datasets, 7 Relation Extraction (RE) datasets, and 6 Event
Extraction (EE) datasets, including three datasets in Chinese. Additionally, to
further improve \KCII's extraction capabilities, particularly its robustness,
event extraction, and multilingual performance, we implemented data augmentation
techniques described below.

For ontology alignment tasks in the knowledge organization phase and tasks in
the knowledge reasoning phase, we are currently generating the corresponding
training data and plan to release the trained models in the future.

\paragraph{Robustness Augmentation.}
To bolster the robustness of \KCII against diverse inputs, we created an
additional training corpus employing augmentation algorithms proposed by
RobustUIE~\citep{zhu2025towards}. Moreover, to efficiently enhance robustness,
we adopted a loss-guided data augmentation approach, selecting 20,000
informative samples from the augmented corpus for training.

\paragraph{Event Augmentation.}
We leveraged the English EE dataset EEMT~\citep{liu2025towards} to strengthen
\KCII's fine-grained event extraction capabilities under extensive schema
scenarios. Furthermore, to boost its Chinese event extraction performance, we
developed a parallel Chinese version of the EEMT dataset.

\paragraph{Multilingual Augmentation.}
Considering the linguistic imbalance between Chinese and English datasets within
the Information Extraction (IE) corpus, we employed the parallel data generation
pipeline introduced by ~\citep{zuo2024alignxie}. Using this method, we
constructed parallel IE corpora from selected English datasets, aiming to
facilitate knowledge sharing across languages and enhance Chinese information
extraction performance.

\KCII is fine-tuned based on the Qwen-2.5-Coder-14B-Instruct
model~\citep{hui2024qwen25codertechnicalreport}, utilizing the LLaMA-Factory
framework~\citep{zheng2024llamafactory}. Parameter-efficient fine-tuning is
achieved through Low-Rank Adaptation
(LoRA)~\citep{hu2021loralowrankadaptationlarge} with a LoRA rank of 32. The
warm-up ratio is set at 0.01, and the learning rate at $3 \times 10^{-4}$.
Sequence length is limited to 2048 tokens, and a batch size of 256 is used.
During inference, the temperature parameter is fixed at 0. All training and
evaluations are performed on eight NVIDIA A100 80GB GPUs.
\section{Experiment}\label{sec:experiment}
\subsection{Tasks and Datasets}
We evaluate \KCII on two kinds of tasks, i.e., knowledge organization and
knowledge reasoning tasks. For the knowledge organization task, We further
evaluate the ontology expansion and knowledge extraction abilities on the
corresponding datasets. For the ontology expansion datasets, we evaluate the
performance on WordNet Sub-taxonomies from~\cite{bansal2014structured}, Graphine
taxonomies from~\cite{liu2021graphine}, and three large-scale taxonomies from
SemEval-2016~\cite{bordea2016semeval} across science, environment, and food
domains. Table~\ref{tab:ontology_datasets} presents the statistics of these
taxonomies, all of which contain entities and definitions curated by human
experts. For all benchmarks, we split the training and tesing dataset following
~\citet{zeng2024codetaxo}. For knowledge extraction, we choose 26 information
extraction benchmarks under the supervised setting. For English IE, we evaluate
performance across 10 benchmarks for NER, 6 benchmarks for RE, 3 benchmarks for
ED, and 2 benchmarks for EAE. For Chinese IE, we evaluate performance on 2
benchmarks for NER, 1 benchmark for RE, 1 benchmark for ED, and 1 benchmark for
EAE. Besides, we also evaluate \KCII under the event extraction benchmark with
massive types EEMT~\cite{liu2025towards} and robustness benchmark
RUIE~\cite{zhu2025towards}. We use abbreviations from P1 to P14 for 14 kinds of
perturbations defined in the RUIE. The detailed benchmark statistics are listed
in Tables~\ref{tab:NER evaluation}, ~\ref{tab:RE_EE evaluation}. For the knowledge reasoning task, we further evalute the
Knowledge Base Question Answer (KBQA) and the report generation abilities. For
the KBQA task, we adopt the most common dataset, WebQSP~\citep{webqsp1}, as the
dataset. For the report generation task, there are no suitable datasets. Thus we
evaluate this abilities on the self-construct dataset about the scientific
report.


\subsection{Baselines}
\paragraph{Knowledge Organization.} 
For the knowledge organization task, we evaluate the performance of \KCII on
several kinds of tasks, including ontology expansion and knowledge extraction including entity extraction,
relation extraction, and event extraction. For the ontology expansion task, we
evaluated the performance of \KCII against several baseline methods, including
TaxoExpan~\citep{shen2020taxoexpan}, STEAM~\citep{yu2020steam},
HEF~\citep{wang2022qen}, Musubu~\citep{takeoka2021low},
TEMP~\citep{liu2021temp}, BoxTaxo~\citep{jiang2023single},
TaxoPrompt~\citep{xu2022taxoprompt}, TaxoInstruct~\citep{shen2024unified}, and
CodeTaxo~\citep{zeng2024codetaxo}. 

For the knowledge extraction task, \KCII is compared with the following
universal IE baselines, including
InstructUIE~\citep{wang2023instructuiemultitaskinstructiontuning},
UniversalNER~\citep{zhou2024universalnertargeteddistillationlarge},
YAYIUIE~\citep{xiao2024yayiuiechatenhancedinstructiontuning},
IEPILE~\citep{gui2024iepileunearthinglargescaleschemabased},
B2NER~\citep{yang2025boundarieslearninguniversalentity},
KnowCoder-V1~\citep{li2024knowcoder}, and KnowCoder-X~\citep{zuo2024alignxie}.
Besides, we also introduce the mainstream LLMs such as Qwen series
LLM~\citep{hui2024qwen25codertechnicalreport}, GPT series
LLM~\citep{achiam2023gpt} in the robustness evaluation. 

\paragraph{Knowledge Reasoning.} 

For the knowledge reasoning phase, we first evaluate the performance of \KCII on
the KBQA (Knowledge Base Question Answering) task. For this task, \KCII is
compared with several baselines, including NSM~\citep{NSM},
TIARA~\citep{grailqa1}, DeCAF~\citep{webqsp1},
KD-CoT~\citep{wang2023knowledgedriven}, StructGPT~\citep{structgpt},
KB-BINDER~\citep{DB-BLINDER}, Cot~\citep{wei2022chain}, ToG~\cite{sun2023think}
and G-Retriever~\cite{he2024g}. Then, we evaluate the performance of the
proposed knowledgeable deep research framework against the following baselines,
i.e., Qwen3 Deep
Research~\footnote{https://chat.qwen.ai/?inputFeature=deep\_research}, Grok3
Deep Research~\footnote{https://grok.com/}, OpenAI Deep
Research~\footnote{https://chatgpt.com/} and WebThinker~\citep{Li2025webthinker}.

\subsection{Results}
\begin{table*}[t]
\centering
\adjustbox{max width=\linewidth}{
\begin{tabular}{lcccccccccc}
\toprule
\multirow{2}{*}{\textbf{Model}} 
& \multicolumn{2}{c}{\textbf{SemEval-Sci}} 
& \multicolumn{2}{c}{\textbf{SemEval-Env}} 
& \multicolumn{2}{c}{\textbf{SemEval-Food}} 
& \multicolumn{2}{c}{\textbf{WordNet}} 
& \multicolumn{2}{c}{\textbf{Graphine}} 
\\
\cmidrule(lr){2-3} \cmidrule(lr){4-5} \cmidrule(lr){6-7} \cmidrule(lr){8-9} \cmidrule(lr){10-11}
& Acc & Wu\&P & Acc & Wu\&P & Acc & Wu\&P & Acc & Wu\&P & Acc & Wu\&P \\
\midrule

\multicolumn{11}{l}{\textbf{Self-supervised Setting}}\\
\rowcolor{teal!20} TaxoExpan & 27.8 & 57.6 & 11.1 & 54.8 & 27.6 & 54.2 & 19.8 & 64.8 & 24.5 & 65.9 \\
\rowcolor{teal!20} STEAM & 36.5 & 68.2 & 36.1 & 69.6 & 34.2 & 67.0 & 23.2 & 62.4 & 20.3 & 63.1 \\
\rowcolor{teal!20} HEF & 53.6 & 75.6 & 55.3 & 71.4 & 47.9 & 73.5 & 16.4 & 60.3 & 25.5 & 66.5 \\
\rowcolor{teal!20} Musubu & 44.9 & 76.2 & 45.3 & 65.4 & 42.3 & 72.4 & 28.5 & 64.0 & 35.4 & 75.2 \\
\rowcolor{teal!20} TEMP & 57.8 & 85.3 & 49.2 & 77.7 & 47.6 & 81.0 & 29.4 & 65.7 & 35.9 & 73.8 \\
\rowcolor{teal!20} BoxTaxo & 31.8 & 64.7 & 38.1 & 75.4 & 31.4 & 66.8 & 26.4 & 63.9 & 29.2 & 68.2 \\
\rowcolor{teal!20} TaxoPrompt & 61.4 & 85.6 & 57.4 & 83.6 & 53.2 & 83.1 & 40.3 & 71.5 & 33.9 & 74.4 \\
\rowcolor{teal!20} TaxoInstruct & 45.9 & 76.2 & 48.8 & 77.2 & 34.3 & 70.2 & 43.3 & 71.8 & 31.8 & 69.0 \\

\multicolumn{11}{l}{\textbf{1-shot Setting}}\\
\rowcolor{red!20} NL (GPT-4o) & 54.8 & 88.3 & 52.5 & 81.3 & 55.5 & 85.6 & 72.2 & 90.7 & 69.8 & 89.1 \\
\rowcolor{red!20} CodeTaxo (GPT-4o) & 67.7 & 89.2 & \textbf{62.5} & 86.1 & 58.1 & 85.3 & 74.5 & 91.3 & 72.9 & 91.0 \\
\rowcolor{red!20} NL (GPT-4o-mini) & 50.0 & 83.0 & 35.0 & 76.1 & 55.1 & 87.2 & 60.1 & 86.0 & 58.3 & 85.2 \\
\rowcolor{red!20} CodeTaxo (GPT-4o-mini) & 58.1 & 85.6 & 42.5 & 76.0 & 55.9 & 85.3 & 68.8 & 89.2 & 61.5 & 85.1 \\
\rowcolor{red!20} KnowCoder-V2  & 59.7 & 80.8 & \textbf{62.5} & \textbf{88.2} & \textbf{60.2} & \textbf{87.6} & \textbf{76.9} & \textbf{91.4} & \textbf{83.3} & \textbf{95.9} \\


\bottomrule
\end{tabular}
}
\caption{Performance on taxonomy expansion across five datasets. Metrics include Accuracy (Acc) and Wu\&Palmer Similarity (Wu\&P).}
\label{tab:taxonomy_colored_blocks}
\end{table*}

\subsubsection{Knowledge Organization}

\paragraph{Results on Ontology Expansion.}
The results for ontology expansion are presented in Table~\ref{tab:taxonomy_colored_blocks}. Compared to self-supervised methods, \KCII outperforms all baselines on three benchmarks in terms of both accuracy and Wu\&Palmer similarity. These results highlight the effectiveness of \KCII's two-stage alignment phase, and show that its code-based definitions significantly enhance the model's ability to analyze the corresponding concepts.

\begin{table*}[htbp]
\centering
\adjustbox{max width=\linewidth}{
\begin{tabular}{c
>{\columncolor[HTML]{FFE9E8}}c 
>{\columncolor[HTML]{FFE9E8}}c 
>{\columncolor[HTML]{FFE9E8}}c 
>{\columncolor[HTML]{FFE9E8}}c 
>{\columncolor[HTML]{FFE9E8}}c 
>{\columncolor[HTML]{FFE9E8}}c 
>{\columncolor[HTML]{FFE9E8}}c 
>{\columncolor[HTML]{FFE9E8}}c 
>{\columncolor[HTML]{FFE9E8}}c 
>{\columncolor[HTML]{FFE9E8}}c 
>{\columncolor[HTML]{FFE9E8}}c 
>{\columncolor[HTML]{FFE9E8}}c
>{\columncolor[HTML]{FFE9E8}}c }
\toprule
\multirow{3}{*}{Model} & \multicolumn{13}{c}{\cellcolor[HTML]{FFE9E8}NER} \\ \cmidrule(l){2-14}
    & \multicolumn{10}{c}{\cellcolor[HTML]{FFE9E8}English} & \multicolumn{2}{c}{\cellcolor[HTML]{FFE9E8}Chinese} & Overall\\ \cmidrule(l){2-11} \cmidrule(l){12-13}\cmidrule(l){14-14}
 & ACE2005 & AnatEM & Conll2003 & BC2GM & BC5CDR & FindVehicle & MultiNerd & MIT Movie & MIT Restaurant & WikiAnn & MSRA & ResumenNER &Avg\\ \midrule
\cellcolor[HTML]{EFEFEF}Bert & 87.3 & 85.8 & 92.4 & 80.9 & 85.3 & 87.1 & 91.3 & 88.8 & 81.0 & 70.6 & 95.0 & 95.9 & 87.1\\ 
\cellcolor[HTML]{EFEFEF}InstructUIE & 79.9 & 85.8 & 91.5 & 80.7 & 89.0 & 87.6 & 90.3 & 89.6 & 82.6 & 64.5 & - & - &-\\ 
\cellcolor[HTML]{EFEFEF}UniversalNER & 86.7 & 88.7 & 93.3 & 82.4 & 88.7 & 98.3 & 93.7 & 90.2 & 82.4 & 84.9 & - & - &-\\
\cellcolor[HTML]{EFEFEF}YAYIUIE & 81.8 & 76.5 & 96.7 & 82.1 & 83.7 & 98.5 & 88.4 & 70.1 & 79.4 &  & 96.0 & -&- \\
\cellcolor[HTML]{EFEFEF}IEPILE & 81.9 & 81.9 & 92.5 & 80.7 & 88.1 & 98.5 & 94.6 & 88.2 & 79.9 & 72.6 & 88.0 & 93.9& 87.0 \\
\cellcolor[HTML]{EFEFEF}B2NER & 83.0 & 89.2 & 92.6 & 82.0 & 88.5 & 97.9 & 94.0 & 90.8 & 83.7 & 85.1 & 92.2 & 95.9 &89.0\\
\cellcolor[HTML]{EFEFEF}KnowCoder-V1 & 86.1 & 86.4 & 94.1 & 82.0 & 89.3 & 99.4 & 96.0 & 90.6 & 81.3 & 87.7 & 40.6 & 16.3 &-\\
\cellcolor[HTML]{EFEFEF}KnowCoder-X & 87.5 & 89.2 & 94.7 & 84.5 & 88.5 & 99.5 & 95.9 & 89.5 & 82.0 & 84.5 & \textbf{96.0} & 96.1 &90.1\\
\cellcolor[HTML]{EFEFEF}KnowCoder-V2 & \textbf{87.8} & \textbf{90.0} & \textbf{95.0} & \textbf{85.9} & \textbf{90.8} & \textbf{99.5} & \textbf{96.0} & \textbf{93.3} & \textbf{82.1} & 86.1 & 94.0 & \textbf{96.2} &\textbf{91.1}\\ \bottomrule
\end{tabular}

}
\caption{The performance of all baselines and our models on NER.}
\label{tab:NER evaluation}
\end{table*}

\begin{table*}[htbp]
\centering
\adjustbox{max width=\linewidth}{
\begin{tabular}{c
>{\columncolor{teal!20}}c 
>{\columncolor{teal!20}}c 
>{\columncolor{teal!20}}c 
>{\columncolor{teal!20}}c 
>{\columncolor{teal!20}}c 
>{\columncolor{teal!20}}c 
>{\columncolor{teal!20}}c 
>{\columncolor{teal!20}}c 
>{\columncolor{red!20}}c 
>{\columncolor{red!20}}c 
>{\columncolor{red!20}}c 
>{\columncolor{red!20}}c 
>{\columncolor{red!20}}c 
>{\columncolor{pink!20}}c 
>{\columncolor{pink!20}}c 
>{\columncolor{pink!20}}c
>{\columncolor{pink!20}}c}
\toprule
 & \multicolumn{8}{c}{\cellcolor{teal!20}RE} & \multicolumn{5}{c}{\cellcolor{red!20}ED} & \multicolumn{4}{c}{\cellcolor{pink!20}EAE} \\ \cmidrule(l){2-18} 
 & \multicolumn{6}{c}{\cellcolor{teal!20}English} & \multicolumn{1}{c}{\cellcolor{teal!20}Chinese} &
 \multicolumn{1}{c}{\cellcolor{teal!20}Overall} &
 \multicolumn{3}{c}{\cellcolor{red!20}English}& \multicolumn{1}{c}{\cellcolor{red!20}Chinese} &
 \multicolumn{1}{c}{\cellcolor{red!20}Overall} &
 \multicolumn{2}{c}{\cellcolor{pink!20}English} & \multicolumn{1}{c}{\cellcolor{pink!20}Chinese} &
 \multicolumn{1}{c}{\cellcolor{pink!20}Overall} \\ \cmidrule(l){2-7}  \cmidrule(l){8-8} \cmidrule(l){9-9} \cmidrule(l){10-12} \cmidrule(l){13-13} \cmidrule(l){14-14} \cmidrule(l){15-16} \cmidrule(l){17-17} \cmidrule(l){18-18} 
\multirow{-3}{*}{Model} & ACE 2005& ADE corpus & Conll 2004 & NYT & SciERC & Semeval RE & DUIE 2.0 & Avg& ACE2005 & CASIE & EEMT & DUEE1.0& Avg & ACE2005 & CASIE & DUEE1.0 & Avg\\ \midrule
\cellcolor[HTML]{EFEFEF}InstructUIE & - & 82.3 & 78.5 & 93.0 & 43.5 & 58.5 & -& - & 43.2 & 67.8 & - & - & -& 56.8 & 63.5 & - & -\\
\cellcolor[HTML]{EFEFEF}YAYIUIE & - & 84.1 & \textbf{79.7} & 90.0 & 40.9 & 61.0 & 81.2 & 72.8& 65.0 & 63.0 & -& 85.0 & - & 62.7 & 64.2 & 78.1 & 68.3 \\
\cellcolor[HTML]{EFEFEF}IEPILE & - & 83.7 & 72.9 & 93.0 & 43.5 & 58.5 & 75.6 & 71.2 & 62.7  & 64.2 & - & 78.1 & - & 63.9 & 56.1 & 75.6& 65.2 \\
\cellcolor[HTML]{EFEFEF}KnowCoder-V1 & 64.5 & 84.3 & 73.3 & 93.7 & 37.4 & 58.5 & 20.4 & 61.7  & 74.2 & 58.2 & 9.1 & 15.9 & 39.4 & 70.3 & 20.8 & 7.9 & 33.0\\
\cellcolor[HTML]{EFEFEF}KnowCoder-X & - & 84.5 & 73.1 & 96.1 & 44.9 & 64.8 & 82.9 & -  & 73.6 & 63.9 & - & \textbf{87.2}&- & 70.0 & 65.0 & \textbf{82.1} & 72.3\\
\cellcolor[HTML]{EFEFEF}KnowCoder-V2 & \textbf{66.3} & \textbf{85.0} & 73.3 & \textbf{96.1} & \textbf{47.3} & \textbf{67.2} & \textbf{83.1}& \textbf{74.0} & \textbf{75.3} & \textbf{67.1} & \textbf{53.5} & 87.1 & \textbf{70.7}&\textbf{71.2} & \textbf{65.2} & 81.4 & 72.6\\ \bottomrule
\end{tabular}

}
\caption{The performance of all baselines and our models on the RE, ED and EAE tasks.}
\label{tab:RE_EE evaluation}
\end{table*}

\paragraph{Results on Knowledge Extraction.}
The results for NER, RE, ED, and EAE tasks are shown in Tables~\ref{tab:NER evaluation}, ~\ref{tab:RE_EE evaluation}, respectively. Compared to \KCI,
\KCII significantly enhances extraction performance across all tasks and
demonstrates strong adaptability to multilingual and multi-event scenarios.
Besides, \KCII outperforms the most advanced information extraction models on
most benchmarks. In English IE, \KCII has achieved a significant average
improvement of 3.03\% and 2.92\% F1 on the NER, RE, ED, and EAE compared to all
baselines.  In Chinese IE, \KCII surpasses the bilingual UIE baselines.
Especially in domain-specific benchmarks such as BC2GM, BC5CDR, and SCIERC,
\KCII surpasses the current state-of-the-art by margins of 1.4\%, 1.5\%, and
4.3\% points, respectively. This highlights \KCII's capability in processing
specialized knowledge, thereby offering compelling support for subsequent tasks.
In the EEMT benchmark, where the type-recalling and partitioned extraction
strategies are removed, \KCII still demonstrates competitive extraction
performance relative to LLM-PEE introduced by~\citep{liu2025towards}.  
Furthermore, compared to \KCI, KnowCoder-X~\citep{zuo2024alignxie}, and LLM-PEE, our
method maintains comparable performance while significantly reducing the prompt
length, thereby enhancing the overall extraction efficiency of the model.
\begin{table*}
\centering
\adjustbox{max width=\linewidth}{
\begin{tabular}{lrrrrrrrrrrrrrrrrrrrrrr}
\toprule
\multirow{2}{*}{\textbf{Model}} & \multicolumn{7}{c}{\textbf{\cellcolor{teal!20} NER}} & \multicolumn{6}{c}{\textbf{\cellcolor{red!20} RE}} & \multicolumn{7}{c}{\textbf{\cellcolor{pink!20} ED}} & \multicolumn{2}{c}{\textbf{\cellcolor{blue!10} Overall}} \\ 
\cmidrule(lr){2-8} \cmidrule(lr){9-14} \cmidrule(lr){15-21} \cmidrule(lr){22-23}
& \cellcolor{teal!20}None & \cellcolor{teal!20}P1 & \cellcolor{teal!20}P2 & \cellcolor{teal!20}P3 & \cellcolor{teal!20}P4 & \cellcolor{teal!20}P5 & \cellcolor{teal!20}$\text{Drop}_\text{avg}$ & \cellcolor{red!20}None & \cellcolor{red!20}P6 &\cellcolor{red!20} P7 & \cellcolor{red!20}P8 & \cellcolor{red!20}P9 & \cellcolor{red!20}$\text{Drop}_\text{avg}$ & \cellcolor{pink!20}None & \cellcolor{pink!20}P10 & \cellcolor{pink!20}P11 & \cellcolor{pink!20}P12 & \cellcolor{pink!20}P13 & \cellcolor{pink!20}P14 & \cellcolor{pink!20}$\text{Drop}_\text{avg}$ & \cellcolor{blue!10}Avg & \cellcolor{blue!10}Rank \\ 
\midrule
\multicolumn{23}{l}{\textbf{Open-source LLMs}}\\
\cellcolor{gray!10} Qwen2.5-14B-Instruct & 
\cellcolor{teal!20}58.6 & \cellcolor{teal!20}53.6 & \cellcolor{teal!20}57.7 & \cellcolor{teal!20}55.0 & \cellcolor{teal!20}56.9 & \cellcolor{teal!20}46.6 & \cellcolor{teal!20}$\text{7.9\%}_\downarrow$ & 
\cellcolor{red!20}22.6 & \cellcolor{red!20}19.2 & \cellcolor{red!20}21.3 & \cellcolor{red!20}17.1 & \cellcolor{red!20}8.8 & \cellcolor{red!20}$\text{26.5\%}_\downarrow$ & 
\cellcolor{pink!20}32.0 & \cellcolor{pink!20}29.8 & \cellcolor{pink!20}30.4 & \cellcolor{pink!20}30.9 & \cellcolor{pink!20}31.4 & \cellcolor{pink!20}32.0  & \cellcolor{pink!20}$\text{3.4\%}_\downarrow$ & 
\cellcolor{blue!10}35.5 & \cellcolor{blue!10}12 \\
\cellcolor{gray!10} Qwen2.5-7B-Instruct & 
\cellcolor{teal!20}53.3 & \cellcolor{teal!20}49.8 & \cellcolor{teal!20}51.2 & \cellcolor{teal!20}50.5 & \cellcolor{teal!20}51.2 & \cellcolor{teal!20}41.3 & \cellcolor{teal!20}$\text{8.4\%}_\downarrow$ & 
\cellcolor{red!20}15.6 & \cellcolor{red!20}13.4 & \cellcolor{red!20}14.0 & \cellcolor{red!20}13.8 & \cellcolor{red!20}3.8 & \cellcolor{red!20}$\text{27.9\%}_\downarrow$ & 
\cellcolor{pink!20}19.0 & \cellcolor{pink!20}18.8 & \cellcolor{pink!20}17.6 & \cellcolor{pink!20}17.9 & \cellcolor{pink!20}18.8 & \cellcolor{pink!20}19.2  & \cellcolor{pink!20}$\text{2.8\%}_\downarrow$ & 
\cellcolor{blue!10}27.6 & \cellcolor{blue!10}14 \\
\cellcolor{gray!10} Qwen2.5-3B-Instruct & 
\cellcolor{teal!20}49.5 & \cellcolor{teal!20}47.5 & \cellcolor{teal!20}46.7 & \cellcolor{teal!20}45.3 & \cellcolor{teal!20}45.5 & \cellcolor{teal!20}40.2 & \cellcolor{teal!20}$\text{9.0\%}_\downarrow$ & 
\cellcolor{red!20}8.9 & \cellcolor{red!20}7.6 & \cellcolor{red!20}8.6 & \cellcolor{red!20}7.4 & \cellcolor{red!20}2.0 & \cellcolor{red!20}$\text{28.1\%}_\downarrow$ & 
\cellcolor{pink!20}13.3 & \cellcolor{pink!20}13.8 & \cellcolor{pink!20}12.3 & \cellcolor{pink!20}12.6 & \cellcolor{pink!20}12.7 & \cellcolor{pink!20}13.6  & \cellcolor{pink!20}$\text{2.3\%}_\downarrow$ & 
\cellcolor{blue!10}22.8 & \cellcolor{blue!10}16 \\
\cellcolor{gray!10} Llama3-8B-Instruct & 
\cellcolor{teal!20}55.4 & \cellcolor{teal!20}52.6 & \cellcolor{teal!20}52.9 & \cellcolor{teal!20}51.1 & \cellcolor{teal!20}53.5 & \cellcolor{teal!20}25.7 & \cellcolor{teal!20}$\text{14.9\%}_\downarrow$ & 
\cellcolor{red!20}17.3 & \cellcolor{red!20}15.0 & \cellcolor{red!20}15.7 & \cellcolor{red!20}13.6 & \cellcolor{red!20}2.5 & \cellcolor{red!20}$\text{32.4\%}_\downarrow$ & 
\cellcolor{pink!20}12.8 & \cellcolor{pink!20}13.1 & \cellcolor{pink!20}13.1 & \cellcolor{pink!20}10.9 & \cellcolor{pink!20}12.5 & \cellcolor{pink!20}12.2 & \cellcolor{pink!20}$\text{3.4\%}_\downarrow$ & 
\cellcolor{blue!10}25.3 & \cellcolor{blue!10}15 \\
\cellcolor{gray!10} Glm-4-9B-Chat & 
\cellcolor{teal!20}57.4 & \cellcolor{teal!20}54.0 & \cellcolor{teal!20}55.8 & \cellcolor{teal!20}51.4 & \cellcolor{teal!20}56.6 & \cellcolor{teal!20}43.2 & \cellcolor{teal!20}$\text{9.0\%}_\downarrow$ & 
\cellcolor{red!20}8.8 & \cellcolor{red!20}7.5 & \cellcolor{red!20}7.5 & \cellcolor{red!20}7.4 & \cellcolor{red!20}1.8 & \cellcolor{red!20}$\text{31.2\%}_\downarrow$ & 
\cellcolor{pink!20}5.6 & \cellcolor{pink!20}6.6 & \cellcolor{pink!20}4.7 & \cellcolor{pink!20}4.1 & \cellcolor{pink!20}5.1 & \cellcolor{pink!20}5.9 & \cellcolor{pink!20}$\text{5.7\%}_\downarrow$ & 
\cellcolor{blue!10}22.6 & \cellcolor{blue!10}17 \\
\cellcolor{gray!10} Internlm2.5-7B-Chat & 
\cellcolor{teal!20}51.6 & \cellcolor{teal!20}48.0 & \cellcolor{teal!20}48.8 & \cellcolor{teal!20}46.9 & \cellcolor{teal!20}45.3 & \cellcolor{teal!20}31.0 & \cellcolor{teal!20}$\text{14.7\%}_\downarrow$ & 
\cellcolor{red!20}12.0 & \cellcolor{red!20}11.3 & \cellcolor{red!20}10.1 & \cellcolor{red!20}9.0 & \cellcolor{red!20}1.7 & \cellcolor{red!20}$\text{33.1\%}_\downarrow$ & 
\cellcolor{pink!20}11.0 & \cellcolor{pink!20}10.3 & \cellcolor{pink!20}10.6 & \cellcolor{pink!20}8.2 & \cellcolor{pink!20}9.6 & \cellcolor{pink!20}11.3 & \cellcolor{pink!20}$\text{9.1\%}_\downarrow$ & 
\cellcolor{blue!10}22.2 & \cellcolor{blue!10}18 \\
\cellcolor{gray!10} CodeLlama-7B-Instruct & 
\cellcolor{teal!20}46.3 & \cellcolor{teal!20}45.0 & \cellcolor{teal!20}45.0 & \cellcolor{teal!20}38.9 & \cellcolor{teal!20}42.4 & \cellcolor{teal!20}14.5 & \cellcolor{teal!20}$\text{19.7\%}_\downarrow$ & 
\cellcolor{red!20}13.7 & \cellcolor{red!20}11.6 & \cellcolor{red!20}12.2 & \cellcolor{red!20}11.3 & \cellcolor{red!20}2.8 & \cellcolor{red!20}$\text{30.8\%}_\downarrow$ & 
\cellcolor{pink!20}8.6 & \cellcolor{pink!20}9.3 & \cellcolor{pink!20}8.8 & \cellcolor{pink!20}6.1 & \cellcolor{pink!20}8.2 & \cellcolor{pink!20}9.2 & \cellcolor{pink!20}$\text{3.3\%}_\downarrow$ & 
\cellcolor{blue!10}19.6 & \cellcolor{blue!10}19 \\
\cellcolor{gray!10} Vicuna-7B-v1.5 & 
\cellcolor{teal!20}39.0 & \cellcolor{teal!20}38.2 & \cellcolor{teal!20}37.4 & \cellcolor{teal!20}35.0 & \cellcolor{teal!20}38.0 & \cellcolor{teal!20}16.7 & \cellcolor{teal!20}$\text{15.2\%}_\downarrow$ & 
\cellcolor{red!20}11.2 & \cellcolor{red!20}11.0 & \cellcolor{red!20}10.1 & \cellcolor{red!20}7.6 & \cellcolor{red!20}0.8 & \cellcolor{red!20}$\text{34.1\%}_\downarrow$ & 
\cellcolor{pink!20}6.9 & \cellcolor{pink!20}7.5 & \cellcolor{pink!20}7.2 & \cellcolor{pink!20}4.5 & \cellcolor{pink!20}6.1 & \cellcolor{pink!20}6.3 & \cellcolor{pink!20}$\text{8.4\%}_\downarrow$ & 
\cellcolor{blue!10}16.7 & \cellcolor{blue!10}20 \\

\multicolumn{23}{l}{\textbf{Closed-source LLMs}}\\
\cellcolor{gray!10} GLM4-Plus & 
\cellcolor{teal!20}63.2 & \cellcolor{teal!20}59.8 & \cellcolor{teal!20}63.0 & \cellcolor{teal!20}61.6 & \cellcolor{teal!20}60.9 & \cellcolor{teal!20}49.7 & \cellcolor{teal!20}$\text{6.6\%}_\downarrow$ & 
\cellcolor{red!20}32.2 & \cellcolor{red!20}29.2 & \cellcolor{red!20}31.3 & \cellcolor{red!20}26.1 & \cellcolor{red!20}5.3 & \cellcolor{red!20}$\text{28.6\%}_\downarrow$ & 
\cellcolor{pink!20}43.5 & \cellcolor{pink!20}39.9 & \cellcolor{pink!20}43.3 & \cellcolor{pink!20}34.6 & \cellcolor{pink!20}40.8 & \cellcolor{pink!20}43.9 & \cellcolor{pink!20}$\text{6.9\%}_\downarrow$ & 
\cellcolor{blue!10}42.8 & \cellcolor{blue!10}9 \\
\cellcolor{gray!10} DeepSeek-V3 & 
\cellcolor{teal!20}62.3 & \cellcolor{teal!20}59.8 & \cellcolor{teal!20}61.5 & \cellcolor{teal!20}61.3 & \cellcolor{teal!20}58.7 & \cellcolor{teal!20}55.0 & \cellcolor{teal!20}$\text{4.9\%}_\downarrow$ & 
\cellcolor{red!20}31.3 & \cellcolor{red!20}29.0 & \cellcolor{red!20}29.6 & \cellcolor{red!20}26.2 & \cellcolor{red!20}10.0 & \cellcolor{red!20}$\text{24.3\%}_\downarrow$ & 
\cellcolor{pink!20}38.8 & \cellcolor{pink!20}37.8 & \cellcolor{pink!20}38.3 & \cellcolor{pink!20}34.5 & \cellcolor{pink!20}35.6 & \cellcolor{pink!20}38.9 & \cellcolor{pink!20}$\text{4.6\%}_\downarrow$  & 
\cellcolor{blue!10}41.7 & \cellcolor{blue!10}10\\
\cellcolor{gray!10} GPT-4-turbo & 
\cellcolor{teal!20}60.6 & \cellcolor{teal!20}57.5 & \cellcolor{teal!20}59.8 & \cellcolor{teal!20}58.2 & \cellcolor{teal!20}56.2 & \cellcolor{teal!20}33.4 & \cellcolor{teal!20}$\text{12.5\%}_\downarrow$ & 
\cellcolor{red!20}33.0 & \cellcolor{red!20}30.0 & \cellcolor{red!20}31.6 & \cellcolor{red!20}26.8 & \cellcolor{red!20}4.5 & \cellcolor{red!20}$\text{29.6\%}_\downarrow$ & 
\cellcolor{pink!20}40.0  & \cellcolor{pink!20}38.0  & \cellcolor{pink!20}39.6  & \cellcolor{pink!20}34.5  & \cellcolor{pink!20}37.3  & \cellcolor{pink!20}39.8  & \cellcolor{pink!20}$\text{5.4\%}_\downarrow$  &
\cellcolor{blue!10}40.0 & \cellcolor{blue!10}11 \\
\cellcolor{gray!10} GPT-3.5-turbo & 
\cellcolor{teal!20}51.8 & \cellcolor{teal!20}47.9 & \cellcolor{teal!20}48.9 & \cellcolor{teal!20}50.5 & \cellcolor{teal!20}39.0 & \cellcolor{teal!20}33.1 & \cellcolor{teal!20}$\text{15.3\%}_\downarrow$ & 
\cellcolor{red!20}23.8 & \cellcolor{red!20}20.6 & \cellcolor{red!20}21.3 & \cellcolor{red!20}16.7 & \cellcolor{red!20}2.4 & \cellcolor{red!20}$\text{35.9\%}_\downarrow$ & 
\cellcolor{pink!20}38.0 & \cellcolor{pink!20}29.5 & \cellcolor{pink!20}36.1 & \cellcolor{pink!20}36.7 & \cellcolor{pink!20}33.9 & \cellcolor{pink!20}36.9 & \cellcolor{pink!20}$\text{8.9\%}_\downarrow$ & 
\cellcolor{blue!10}33.3 & \cellcolor{blue!10}13 \\
\midrule
\multicolumn{23}{l}{\textbf{Traditional IE Models}}\\
\cellcolor{gray!10} Stanza & 
\cellcolor{teal!20}80.7 & \cellcolor{teal!20}70.1 & \cellcolor{teal!20}77.1 & \cellcolor{teal!20}71.5 & \cellcolor{teal!20}78.1 & \cellcolor{teal!20}51.1 & \cellcolor{teal!20}$\text{13.8\%}_\downarrow$ & 
\cellcolor{red!20}- & \cellcolor{red!20}- & \cellcolor{red!20}- & \cellcolor{red!20}- & \cellcolor{red!20}- & \cellcolor{red!20}- & 
\cellcolor{pink!20}-  & \cellcolor{pink!20}-  & \cellcolor{pink!20}-  & \cellcolor{pink!20}-  & \cellcolor{pink!20}-  & \cellcolor{pink!20}-  & \cellcolor{pink!20}-  &
\cellcolor{blue!10}- & \cellcolor{blue!10}-\\
\cellcolor{gray!10} TNER & 
\cellcolor{teal!20}83.0 & \cellcolor{teal!20}73.3 & \cellcolor{teal!20}78.0 & \cellcolor{teal!20}73.9 & \cellcolor{teal!20}81.0 & \cellcolor{teal!20}73.2 & \cellcolor{teal!20}$\text{8.6\%}_\downarrow$ & 
\cellcolor{red!20}- & \cellcolor{red!20}- & \cellcolor{red!20}- & \cellcolor{red!20}- & \cellcolor{red!20}- & \cellcolor{red!20}- & 
\cellcolor{pink!20}-  & \cellcolor{pink!20}-  & \cellcolor{pink!20}-  & \cellcolor{pink!20}-  & \cellcolor{pink!20}-  & \cellcolor{pink!20}-  & \cellcolor{pink!20}-  &
\cellcolor{blue!10}- & \cellcolor{blue!10}- \\
\cellcolor{gray!10} PFN & 
\cellcolor{teal!20}- & \cellcolor{teal!20}- & \cellcolor{teal!20}- & \cellcolor{teal!20}- & \cellcolor{teal!20}- & \cellcolor{teal!20}- & \cellcolor{teal!20}- & 
\cellcolor{red!20}76.3 & \cellcolor{red!20}58.6 & \cellcolor{red!20}73.8 & \cellcolor{red!20}68.4 & \cellcolor{red!20}20.4 & \cellcolor{red!20}$\text{27.5\%}_\downarrow$  & 
\cellcolor{pink!20}-  & \cellcolor{pink!20}-  & \cellcolor{pink!20}-  & \cellcolor{pink!20}-  & \cellcolor{pink!20}-  & \cellcolor{pink!20}-  & \cellcolor{pink!20}-  & 
\cellcolor{blue!10}- & \cellcolor{blue!10}-  \\
\cellcolor{gray!10} EEQA & 
\cellcolor{teal!20}- & \cellcolor{teal!20}- & \cellcolor{teal!20}- & \cellcolor{teal!20}- & \cellcolor{teal!20}- & \cellcolor{teal!20}- & \cellcolor{teal!20}- & 
\cellcolor{red!20}- & \cellcolor{red!20}- & \cellcolor{red!20}- & \cellcolor{red!20}- & \cellcolor{red!20}- & \cellcolor{red!20}- &
\cellcolor{pink!20}68.3  & \cellcolor{pink!20}53.2  & \cellcolor{pink!20}64.0  & \cellcolor{pink!20}63.4  & \cellcolor{pink!20}59.7  & \cellcolor{pink!20}66.5  & \cellcolor{pink!20}$\text{10.1\%}_\downarrow$ & 
\cellcolor{blue!10}- & \cellcolor{blue!10}-  \\
\multicolumn{23}{l}{\textbf{UIE Models}}\\
\cellcolor{gray!10} UIE & 
\cellcolor{teal!20}83.9 & \cellcolor{teal!20}74.3 & \cellcolor{teal!20}81.1 & \cellcolor{teal!20}75.6 & \cellcolor{teal!20}81.7 & \cellcolor{teal!20}70.3 & \cellcolor{teal!20}$\text{8.7\%}_\downarrow$ & 
\cellcolor{red!20}84.1 & \cellcolor{red!20}63.3 & \cellcolor{red!20}81.1 & \cellcolor{red!20}77.0 & \cellcolor{red!20}35.9 & \cellcolor{red!20}$\text{23.5\%}_\downarrow$ & 
\cellcolor{pink!20}71.3 & \cellcolor{pink!20}52.6 & \cellcolor{pink!20}65.2 & \cellcolor{pink!20}67.7 & \cellcolor{pink!20}67.6 & \cellcolor{pink!20}68.8 & \cellcolor{pink!20}$\text{9.9\%}_\downarrow$ & 
\cellcolor{blue!10}70.7 & \cellcolor{blue!10}6\\
\cellcolor{gray!10} InstructUIE-11B & 
\cellcolor{teal!20}73.9 & \cellcolor{teal!20}65.7 & \cellcolor{teal!20}69.5 & \cellcolor{teal!20}64.3 & \cellcolor{teal!20}72.0 & \cellcolor{teal!20}70.3 & \cellcolor{teal!20}$\text{7.5\%}_\downarrow$ & 
\cellcolor{red!20}68.4 & \cellcolor{red!20}48.3 & \cellcolor{red!20}66.3 & \cellcolor{red!20}61.6 & \cellcolor{red!20}56.1 & \cellcolor{red!20}$\text{15.1\%}_\downarrow$  & 
\cellcolor{pink!20}60.6  & \cellcolor{pink!20}50.8  & \cellcolor{pink!20}56.8  & \cellcolor{pink!20}59.5  & \cellcolor{pink!20}59.3  & \cellcolor{pink!20}59.6  & \cellcolor{pink!20}$\text{5.6\%}_\downarrow$  & 
\cellcolor{blue!10}62.5 & \cellcolor{blue!10}7  \\
\cellcolor{gray!10} YAYI-UIE-13B & 
\cellcolor{teal!20}80.7 & \cellcolor{teal!20}69.3 & \cellcolor{teal!20}75.3 & \cellcolor{teal!20}72.6 & \cellcolor{teal!20}79.2 & \cellcolor{teal!20}75.6 & \cellcolor{teal!20}$\text{7.8\%}_\downarrow$ & 
\cellcolor{red!20}66.4 & \cellcolor{red!20}47.3 & \cellcolor{red!20}65.0 & \cellcolor{red!20}59.9 & \cellcolor{red!20}38.4 & \cellcolor{red!20}$\text{20.7\%}_\downarrow$  & 
\cellcolor{pink!20}42.7  & \cellcolor{pink!20}37.5  & \cellcolor{pink!20}41.3  & \cellcolor{pink!20}41.4  & \cellcolor{pink!20}40.8  & \cellcolor{pink!20}41.6  & \cellcolor{pink!20}$\text{5.2\%}_\downarrow$  & 
\cellcolor{blue!10}57.4 & \cellcolor{blue!10}8  \\
\cellcolor{gray!10} KnowCoder-7B & 
\cellcolor{teal!20}\textbf{87.4} & \cellcolor{teal!20}76.4 & \cellcolor{teal!20}81.3 & \cellcolor{teal!20}79.6 & \cellcolor{teal!20}84.7 & \cellcolor{teal!20}81.5 & \cellcolor{teal!20}$\text{7.7\%}_\downarrow$ & 
\cellcolor{red!20}84.0 & \cellcolor{red!20}57.3 & \cellcolor{red!20}80.5 & \cellcolor{red!20}76.4 & \cellcolor{red!20}73.3 & \cellcolor{red!20}$\text{14.4\%}_\downarrow$  & 
\cellcolor{pink!20}74.2  & \cellcolor{pink!20}53.8  & \cellcolor{pink!20}69.6  & \cellcolor{pink!20}70.5 & \cellcolor{pink!20}70.3  & \cellcolor{pink!20}\textbf{72.4}  & \cellcolor{pink!20}$\text{9.3\%}_\downarrow$  & 
\cellcolor{blue!10}74.9 & \cellcolor{blue!10}3  \\
\cellcolor{gray!10} $\text{KnowCoder-7B}_\text{partial}$ & 
\cellcolor{teal!20}84.4 & \cellcolor{teal!20}73.8 & \cellcolor{teal!20}80.1 & \cellcolor{teal!20}81.1 & \cellcolor{teal!20}82.1 & \cellcolor{teal!20}79.0 & \cellcolor{teal!20}$\text{6.1\%}_\downarrow$ & 
\cellcolor{red!20}81.4 & \cellcolor{red!20}60.6 & \cellcolor{red!20}79.1 & \cellcolor{red!20}74.5 & \cellcolor{red!20}52.8 & \cellcolor{red!20}$\text{18.0\%}_\downarrow$  & 
\cellcolor{pink!20}69.1  & \cellcolor{pink!20}55.5  & \cellcolor{pink!20}66.1  & \cellcolor{pink!20}65.8  & \cellcolor{pink!20}66.8  & \cellcolor{pink!20}67.8  & \cellcolor{pink!20}$\text{6.8\%}_\downarrow$  & 
\cellcolor{blue!10}71.8 & \cellcolor{blue!10}5  \\

\cellcolor{gray!10} $\text{KnowCoder-7B-Robust}$ & 
\cellcolor{teal!20}85.9 & \cellcolor{teal!20}81.3 & \cellcolor{teal!20}83.5 & \cellcolor{teal!20}86.4 & \cellcolor{teal!20}\textbf{86.1} & \cellcolor{teal!20}\textbf{84.6} & \cellcolor{teal!20}\textbf{$\text{1.7\%}_\downarrow$} & 
\cellcolor{red!20}83.1 & \cellcolor{red!20}66.0 & \cellcolor{red!20}82.9 & \cellcolor{red!20}81.1 & \cellcolor{red!20}79.8  & \cellcolor{red!20}$\text{6.8\%}_\downarrow$  & 
\cellcolor{pink!20}70.2 & \cellcolor{pink!20}65.7  & \cellcolor{pink!20}67.1  & \cellcolor{pink!20}70.5  & \cellcolor{pink!20}68.3  & \cellcolor{pink!20}69.6  & \cellcolor{pink!20}$\text{2.8\%}_\downarrow$ & 
\cellcolor{blue!10}77.2 & \cellcolor{blue!10}2  \\
\cellcolor{gray!10} $\text{KnowCoder-7B-Robust}_{\text{LDA}}$ & 
\cellcolor{teal!20}86.1 & \cellcolor{teal!20}81.2 & \cellcolor{teal!20}\textbf{84.9} & \cellcolor{teal!20}86.5 & \cellcolor{teal!20}85.6 & \cellcolor{teal!20}83.8 & \cellcolor{teal!20}$\text{1.9\%}_\downarrow$ & 
\cellcolor{red!20}82.2 & \cellcolor{red!20}\textbf{66.5} & \cellcolor{red!20}82.5 & \cellcolor{red!20}81.3 & \cellcolor{red!20}\textbf{81.3} & \cellcolor{red!20}\textbf{$\text{5.2\%}_\downarrow$}  & 
\cellcolor{pink!20}69.4  & \cellcolor{pink!20}66.0  & \cellcolor{pink!20}68.1  & \cellcolor{pink!20}70.0  & \cellcolor{pink!20}68.8  & \cellcolor{pink!20}68.9  & \cellcolor{pink!20}\textbf{$\text{1.5\%}_\downarrow$} & 
\cellcolor{blue!10}77.2 & \cellcolor{blue!10}2  \\
\midrule
\cellcolor{gray!10} \KCII & 
\cellcolor{teal!20}\textbf{87.4} & \cellcolor{teal!20}\textbf{81.7} & \cellcolor{teal!20}83.9 & \cellcolor{teal!20}\textbf{87.9} & \cellcolor{teal!20}84.9 & \cellcolor{teal!20}84.0 & \cellcolor{teal!20}$\text{2.4\%}_\downarrow$ & 
\cellcolor{red!20}84.6 & \cellcolor{red!20}\textbf{65.8} & \cellcolor{red!20}87.3 & \cellcolor{red!20}\textbf{81.9} & \cellcolor{red!20}78.8 & \cellcolor{red!20}$\text{5.6\%}_\downarrow$  & 
\cellcolor{pink!20}\textbf{74.7}  & \cellcolor{pink!20}\textbf{74.6}  & \cellcolor{pink!20}\textbf{70.8}  & \cellcolor{pink!20}\textbf{72.9}  & \cellcolor{pink!20}\textbf{72.9}  & \cellcolor{pink!20}69.5  & \cellcolor{pink!20}$\text{3.7\%}_\downarrow$  & 
\cellcolor{blue!10}\textbf{78.6} & \cellcolor{blue!10}1  \\
\bottomrule
\end{tabular}
}
\caption{The performance of all baselines and our models on RUIE-bench.}
\label{tab:robust evaluation}
\end{table*}
The results for NER, RE, and ED tasks under robustness settings are presented in
Table~\ref{tab:robust evaluation}. Compared to \KCI and KnowCoder-7B-Robust,
\KCII demonstrates strong robustness across various perturbations and ranks 1st
among all baselines with average of 78.6\%, surpass the KnowCoder-7B-Robust by
1.4\%. Notably, under more complex robustness setting such as Extended Sentence
perturbation, \KCII consistently maintains and or even improves its performance
than None setting across all tasks (87.3\% in P7 v.s 84.6\% in None). This
indicates that \KCII is capable of handling longer and more complex texts while
still delivering accurate knowledge organization.

\subsubsection{Knowledge Reasoning.}

\begin{wraptable}{r}{0.35\textwidth}
    \centering
    \resizebox{0.9\linewidth}{!}{
    \begin{tabular}{lc}
    \toprule
    \textbf{Method} & \textbf{WebQSP} \\ \midrule
    \multicolumn{2}{c}{\textit{Fine-tuned}} \\ \midrule
    NSM \citep{NSM}                   & 74.3 \\
    TIARA \citep{grailqa1}            & 75.2 \\
    DeCAF \citep{webqsp1}             & 82.1 \\ \midrule
    \multicolumn{2}{c}{\textit{Prompting}} \\ \midrule
    KD-CoT \citep{wang2023knowledgedriven} & 73.7 \\
    StructGPT \citep{structgpt}       & 72.6 \\
    KB-BINDER \citep{DB-BLINDER}      & 74.4 \\ 
    CoT w. LLama2-70B-Chat             & 57.4 \\
    ToG-R w. LLama2-70B-Chat               & 68.9 \\
    ToG  w. LLama2-70B-Chat              & 63.7 \\
    CoT w. GPT-4                               & 67.3 \\
    ToG-R  w. GPT-4                               & 81.9 \\
    ToG   w. GPT-4                                & 82.6 \\
    G-Retriever                                   & 70.1 \\
    EtD w. ChatGPT                                & 82.5 \\
    \midrule       
   \KCII                              & \textbf{84.7} \\
    \bottomrule    
    \end{tabular}
    }
    \caption{Performances of \KCII on the KBQA task.}
\label{table: kbqa}
\end{wraptable}
\paragraph{Results on Knowledge Base Question Answering.}
Following ToG~\citep{sun2023think}, we utilize the ground truth topic entity for
each question. By centering on this topic entity, we extract the corresponding
subgraph for each question. After converting the subgraph into class definition
and instance code, \KCII generates the analysis code based on the given
questions. The Hits@1 scores for the KBQA task on WebQSP~\citep{webqsp1} are shown
in Table~\ref{table: kbqa}. Compared to the fine-tuned baseline, \KCII
achieves a 2.6\% improvement, even without any fine-tuning on the corresponding
dataset. Additionally, when compared to prompting-based methods, \KCII shows a
2.2\% improvement. These results suggest that \KCII is capable of generating
accurate analysis code for the KBQA task.

\begin{table}[t]
   \centering
   \adjustbox{max width=\linewidth}{
   \begin{tabular}{lcccccc}
   \toprule
   \multicolumn{7}{c}{\textbf{Report Generation Task}} \\
   \midrule
   \textbf{Method} & Comp. & Thorough. & Fact. & Coherence & Insight & Avg. \\
   \midrule

   \multicolumn{6}{l}{\textit{Closed-source System}} \\
   
   Qwen3 DeepResearch   & 6.9 &	6.4&	6.7&	6.7&	4.3&	6.2 \\
   Grok3 DeeperSearch   & 6.8	&6.4	&7.8	&7.8	&5.3	&6.8 \\
   Openai DeepResearch   & 8.3 &	8.5&	8.3&	8.7&	4.8&	7.7 \\
   \midrule
   \multicolumn{6}{l}{\textit{Open-source System}} \\
    WebThinker & 7.3 &	6.4&	6.9&	7.4&	4.1&	6.4 \\
   \rowcolor{purple!10} KnowCoder-V2  & \textbf{9.3}&	\textbf{9.2}&	\textbf{8.5}&	\textbf{8.8}&\textbf{8.7}&	\textbf{8.9}
\\
   \bottomrule
   \end{tabular}
   }
   \caption{Performance on the report generation task.}
   \label{tab:report}
   \end{table}

\paragraph{Results on Report Generation.}
To evaluate the report generation ability of the proposed knowledgeable deep
research framework empowered by \KCII, we construct a self-constructed dataset
about the scientific report. The dataset focuses on researching the academic
achievements about scientists and generate their corresponding reports. 
Following the setting in~\cite{Li2025webthinker}, we use LLMs to give scores to
reports generated by different methods from several aspects, including the
completeness, thoroughness, factuality, coherence and an additional score called
``insight''. The scores are averaged from two LLMs, including DeepSeek-V3 and
DeepSeek-R1. The results are shown in Table~\ref{tab:report}. \KCII achieves
the highest score among all baselines on all metrics. Due the knowledge
computation steps, \KCII is able to generate more comprehensive and insightful
experimental results and more deep analysis, leading to the significant
improvement on ``Insigt'' score.

\begin{figure}
    \centering
    \includegraphics[width=\textwidth]{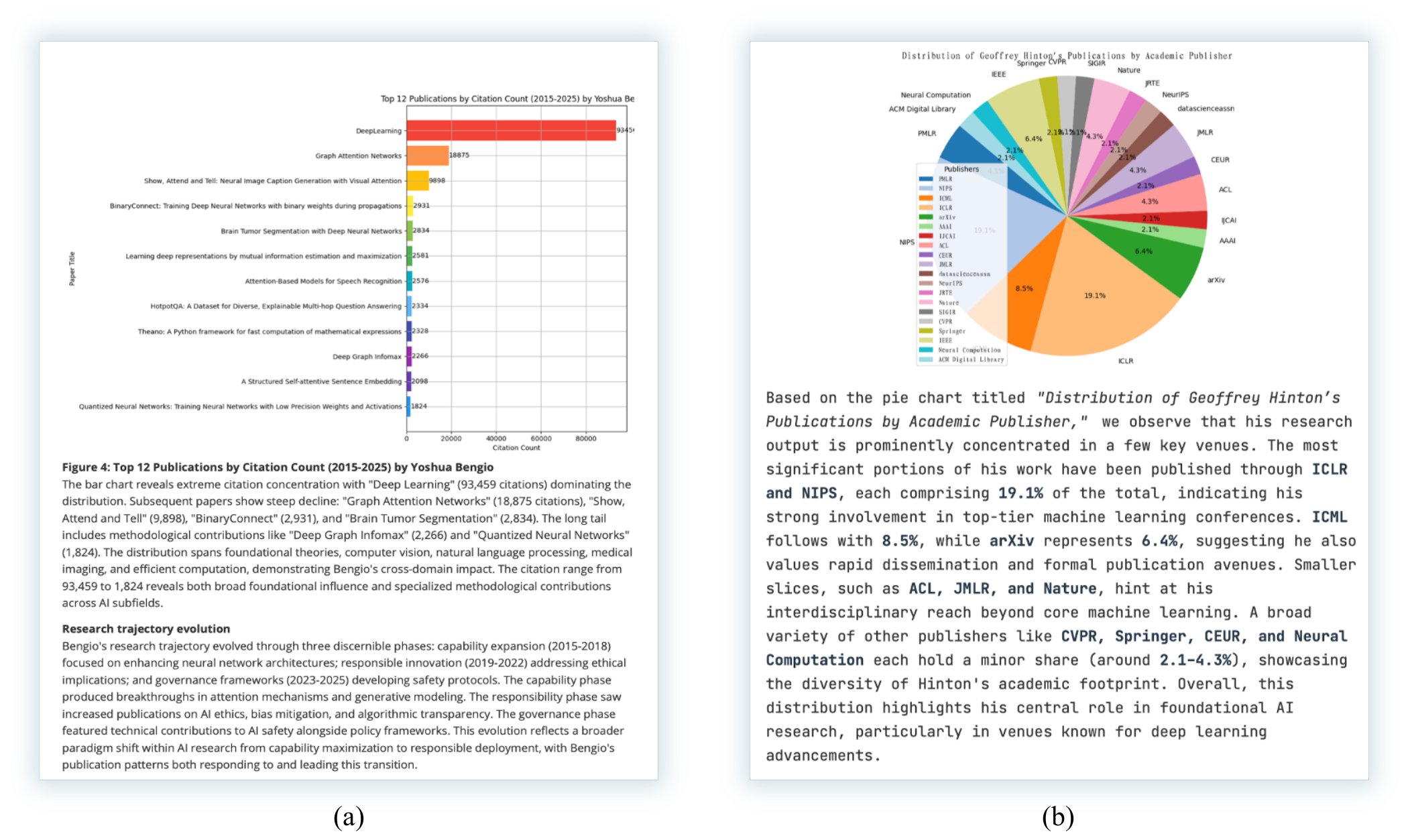}
    \caption{Two cases of reports generated by the KDR framework.}
    \label{fig:my_label}
\end{figure}

\paragraph{Case Study.} To illustrate the depth of knowledge analysis that KDR can produce, we present two cases of partial reports generated by \KCII under the KDR framework, based on the following queries:
(1) "Search and summarize Geoffrey Hinton's research achievements in papers over the past decade (2015 to 2025), and generate a report," and (2) "Search and summarize Yoshua Bengio's research achievements in papers over the past decade (2015 to 2025), and generate a report." Based on the organized structured knowledge, the KDR framework automatically decomposes the research task and designs relevant experiments for \KCII. Leveraging the knowledge computation cycle, \KCII can compute citation counts for Bengio’s publications and visualize them in a chart. For Hinton, it can automatically analyze his publication preferences over the past decade and present the results in a pie chart. By analyzing these experimental results, KDR can identify each scientist's most influential academic contributions and their preferred publication venues. These analyses can offer novel, insightful findings rather than simply summarizing information already available online.


\section{Conclusion}\label{sec:conclusion} In this paper, we introduced the
Knowledgeable Deep Research framework (KDR), which enhances traditional deep
research frameworks by integrating advanced knowledge analysis capabilities. The
framework features an independent knowledge organization phase that preprocesses
large-scale, domain-specific data into systematic knowledge offline. This
knowledge serves as the foundation for the framework’s online reasoning steps,
enabling complex knowledge computations. We also presented \KCII, an LLM
designed to perform deep knowledge analysis tasks through unified code
generation. For knowledge organization tasks, \KCII generates instantiation code
to transform raw data into structured knowledge objects. For knowledge
computation tasks, it generates analysis code that processes these objects to
derive deep insights. Our extensive experiments across more than thirty
benchmarks, spanning tasks such as ontology expansion, knowledge extraction, and
knowledge base question answering, demonstrate the effectiveness of \KCII.
Through \KCII, the KDR framework is capable of producing high-quality, in-depth
reports with comprehensive analysis.

\bibliography{reference}
\bibliographystyle{reference}

\end{document}